\documentclass[sigconf]{acmart}

\usepackage{amsmath} 
\usepackage{amssymb}  
\usepackage{algorithm}
\usepackage{algorithmicx}
\usepackage{algpseudocode}
\usepackage{xcolor}
\usepackage{colortbl}
\usepackage{graphicx}
\usepackage{newfloat}
\usepackage{listings}
\usepackage{booktabs}
\usepackage{multirow}
\usepackage{makecell}
\usepackage{subcaption}
\usepackage{cleveref}
\usepackage[table]{xcolor}

\AtBeginDocument{%
  }

\setcopyright{acmlicensed}
\copyrightyear{2026}
\acmYear{2026}
\acmDOI{XXXXXXX.XXXXXXX}
\acmConference[MM '26]{the 34th ACM International Conference on Multimedia}{November 10--14,
   2026}{Rio de Janeiro, Brazil}
\acmISBN{978-1-4503-XXXX-X/2018/06}

\acmSubmissionID{3702}

\begin{document}

\title{AdvNav: Behavior-Guided Black-Box Adversarial Attacks on Vision-Language Navigation}

\author{Chenyang Li}
\affiliation{%
  \institution{The Hong Kong University of Science and Technology (Guangzhou)}
  \city{Guang Zhou}
  \country{China}}
\email{lichenyang20020820@gmail.com}

\author{Kaige Li}
\affiliation{%
  \institution{Sun Yat-sen University}
  \city{ShenZhen}
  \country{China}}
\email{likg@mail.sysu.edu.cn}

\author{Zeyu Jiang}
\affiliation{%
  \institution{The Hong Kong University of Science and Technology (Guangzhou)}
  \city{Guang Zhou}
  \country{China}}
\email{zjiang739@gmail.com}

\author{Changhao Chen}
\authornote{Corresponding author.}
\affiliation{%
  \institution{The Hong Kong University of Science and Technology (Guangzhou)}
  \city{GuangZhou}
  \country{China}}
\email{changhaochen@hkust-gz.edu.cn}


\begin{abstract}
 Despite progress in Embodied AI, Vision-and-Language Navigation (VLN) systems remain vulnerable to adversarial visual disturbances. Most existing methods rely on white-box access to target model gradients, which is often unrealistic for real-world deployed systems and computationally exhaustive due to recursive backpropagation for optimization, limiting their applicability. While previous black-box methods predominantly target single-step, instantaneous decision tasks, they struggle to handle the task complexities and temporal dependencies of VLN. This highlights the need for a gradient-free attack method that can effectively disrupt the multi-step sequential perception–action loop using only observable inputs and outputs. Therefore, we propose AdvNav, a behavior-guided black-box adversarial attack framework that disturbs an agent’s first-person views during navigation. To construct an informative surrogate objective for effective optimization guidance in gradient-free search under the black-box setting, we design a dual-granularity behavior-based feedback, aggregating a trajectory-level performance score representing overall navigation degradation, an action-level reward score considering the potential decision risk, and a deviation indicator, all of which are extracted from the agent's self-output behaviors. This feedback guides a hybrid optimization strategy that (i) heuristically tunes perturbation strength via adaptive updates and (ii) evolves noise spatial structure genetically, to iteratively discover the most disruptive noise configuration. Evaluated against Transformer-based model HAMT and LLM-based model MapGPT with two types of backbones on R2R dataset, AdvNav achieves 49.70\%, 65.96\%, and 87.30\% Attack Success Rate, respectively. The result demonstrates the effectiveness and generality of AdvNav, reveals critical perception vulnerabilities and offers insights for the design of future resilient VLN models.
\end{abstract}




\maketitle
\section{Introduction}

Recent advances in Embodied AI have significantly enhanced agents’ capabilities in perception, reasoning, and action~\cite{liu2025embodiedai, wang2022embodiednavigation}. Among these, Vision-and-Language Navigation (VLN)~\cite{anderson2018vlnr2r} has emerged as a fundamental task, requiring agents to follow natural language instructions to reach specified targets in unseen complex environments.
VLN extends intelligence from passive scene understanding to active, goal-driven exploration, with broad applications in interactive multimedia~\cite{chi2020justask}, multimodal AI systems~\cite{driess2023palm}, and mobile service robotics across diverse domains such as medical support~\cite{fiske2019medicalembodiedai}, educational guidance~\cite{memarian2024educationembodiedai}, and industrial assistance~\cite{ren2024industryembodiedai}.

\begin{figure}[t]
    \centering
    \includegraphics[width=\linewidth]{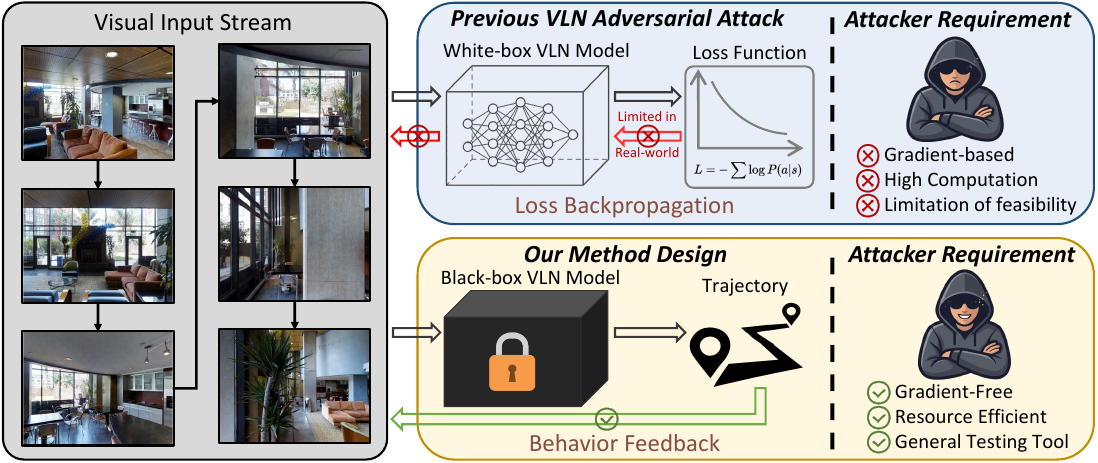}
    \caption{Comparison between existing and our target adversarial settings for VLN. 
    Most prior vision-based adversarial attacks assume a white-box setting requiring access to model gradients, while real-world deployed systems are typically evaluated under black-box constraints with limited computational resources.
    \textit{Our goal is to develop a gradient-free and cost-efficient framework for VLN robustness evaluation.}}
    \label{fig:concept}
\end{figure}

Despite substantial progress in performance~\cite{wu2024embodiednavigationsurvey, zhang2024vlnsurvey}, most VLN systems rely on end-to-end deep neural networks (DNNs) for perception and decision-making
, while DNNs are well known to be vulnerable to adversarial attacks~\cite{carlini2017attackdnncw}.
Consequently, the robustness of deployed VLN agents under realistic conditions remains uncertain.
In practice, these agents often operate in noisy, dynamic, and partially observable environments, while their internal states are often inaccessible due to proprietary constraints. This raises critical concerns for safety-critical applications—for example, in factories, even minor navigation errors, such as trajectory deviations or unexpected stops, can disrupt production, damage equipment, or endanger human workers. This motivates a key question: \textit{how vulnerable are VLN agents to adversarial perturbations when we don't have any knowledge of their internals?}

Adversarial attacks have been extensively investigated in the computer vision community, demonstrating that DNNs can be highly sensitive to small, carefully crafted input perturbations. Gradient-based methods, including FGSM~\cite{goodfellow2014imagefgsm}, PGD~\cite{madry2017pgd}, and CW~\cite{carlini2017attackdnncw}, as well as physically realizable adversarial patches~\cite{brown2017patchimageclassfication}, have been widely applied to tasks such as image classification~\cite{brown2017patchimageclassfication}, object detection~\cite{liu2018patchobjectdetection}, traffic sign recognition~\cite{eykholt2018patchtrafficsign}, and face recognition~\cite{sharif2016patchfacerecoginition}. These studies have largely advanced the understanding of vulnerabilities in deep learning systems.
More recently, adversarial attacks have been extended to embodied navigation. Existing approaches targeting an agent’s visual perception can be broadly categorized into two groups: (1) scene-level perturbations, which manipulate the environment by altering object textures or geometry~\cite{liu2020spatiotemporal, yang2025hijacking} or injecting adversarial patches~\cite{chen2024physicalizable} to mislead perception and planning; and (2) agent-view perturbations, which directly modify the agent’s visual input stream~\cite{ying2023uap} to induce trajectory deviation. However, most prior work assumes a white-box setting, requiring access to model gradients for perturbation optimization~\cite{chakraborty2018adversarial}. This assumption is often impractical in real-world deployments, where models are proprietary and internal information is inaccessible. Consequently, white-box attacks are limited in applicability, tend to be architecture-dependent, and incur high computational cost due to long-horizon gradient-based optimization.

These limitations motivate the need for black-box formulations that assess model robustness using only observable inputs and outputs~\cite{chen2017zoo,ilyas2018nes}. Black-box methods are gradient-free and generally more applicable across different architectures. Representative approaches such as ZOO~\cite{chen2017zoo}, NES~\cite{ilyas2018nes}, and SimBA~\cite{guo2019simba} have demonstrated effectiveness in vision tasks~\cite{bhambri2019blackboxCV}. However, these methods are primarily designed for single-step decision problems, where optimization benefits from immediate and dense feedback signals.
In contrast, it is challenging to black-box adversarial optimization on VLN, which is inherently a multi-step, multimodal sequential decision-making task, due to its delayed feedback, non-linear error accumulation, and the agent's inherent capacity for policy self-correction. Therefore, the attack must account for long-term behavioral effects rather than instantaneous outputs.

In this work, we propose \textbf{AdvNav}, a novel behavior-based black-box \textbf{Adv}ersarial attacks framework for vision-language \textbf{Nav}igation models. 
AdvNav injects spatially coherent perturbations into the agent’s first-person visual observations throughout the navigation process, inducing continuous disruptions in perception. 
To overcome the absence of gradient information in the black-box setting, we introduce a dual-granularity behavior-based feedback, providing a surrogate signal derived solely from the target model's observable behaviors to guide the perturbation optimization. 
This feedback combines: (1) a trajectory-level performance score, capturing overall navigation disruption; (2) an action-level reward score, sensitive to per-step action drift and addressing metric sparsity when the agent appears to follow the reference trajectory but latent errors exist; and (3) an indicator signaling trajectory deviation. Built upon this feedback, we design a hybrid optimization strategy that couples perturbation intensity tuning via adaptive updates with noise structure refinement through genetic evolution, enabling efficient exploration of highly disruptive perturbations.
Extensive experiments on both Transformer-based and LLM-based VLN models demonstrate the effectiveness and generality of AdvNav. On the R2R dataset, AdvNav achieves a 49.70\% Attack Success Rate (ASR) on the Transformer-based HAMT~\cite{chen2021hamt}, and 65.96\% and 87.30\% on MapGPT~\cite{chen2024mapgpt} across Qwen3-VL and GPT-4V backbones, respectively, illustrating its efficacy and broad applicability as a robustness evaluation tool for existing VLN agents.

In summary, our contributions are as follows:
\begin{itemize}
    \item We propose AdvNav, a general behavior-guided black-box adversarial attack framework targeting visual perception of VLN agents, providing an effective and cost-efficient tool for robustness evaluation.
    \item We develop a gradient-free hybrid perturbation optimization strategy that combines intensity adaptive tuning with structure refinement via genetic evolution, leveraging dual-granularity behavior feedback tailored for VLN tasks.
    \item We conduct extensive experiments on Transformer- and LLM-based VLN models across 11 indoor scenes with multiple language instructions, demonstrating that AdvNav significantly impacts navigation performance and exposes the vulnerability of current VLN systems. 
\end{itemize}

\section{Related Work}

\subsection{Visual-and-Language Navigation (VLN)}
Embodied navigation tasks~\cite{wang2022embodiednavigation} require agents to interpret visual input and interact with previously unseen environments to accomplish specific goals. These tasks can be categorized by target type, including Object Goal Navigation~\cite{chaplot2020objectgoal}, Image Goal Navigation~\cite{zhu2017imagegoal}, Embodied Question Answering (EQA)~\cite{das2018eqa}, and Vision-and-Language Navigation (VLN)~\cite{anderson2018vlnr2r}. Among them, VLN is particularly challenging, as agents must follow high-level natural language instructions to reach a described location, requiring both semantic understanding and alignment between linguistic commands and visual observations. 
Approaches for these tasks have progressed from learning-based methods~\cite{anderson2018vlnr2r, wang2019reinforced} focusing on end-to-end training, to pretrained Transformer-based models~\cite{chen2021hamt, chen2022duet, qiao2022hop} with stronger capability on cross-model alignment, and LLM-based models~\cite{zhou2024navgpt, chen2024mapgpt, zheng2024towardsllmmodel} that leverage large-scale knowledge for reasoning. 
In this work, we choose a representative Transformer-based model HAMT~\cite{chen2021hamt}, and a zero-shot LLM-based model MapGPT~\cite{chen2024mapgpt}, to better investigate the visual robustness and understand perception vulnerabilities of these advanced VLN systems.

\subsection{Adversarial Attacks}
Adversarial attacks expose the fragility of deep neural networks by introducing human-imperceptible perturbations that mislead predictions. Early gradient-based methods such as FGSM~\cite{goodfellow2014imagefgsm}, PGD~\cite{madry2017pgd}, and CW~\cite{carlini2017attackdnncw} remain foundational and have been validated across visual recognition tasks~\cite{brown2017patchimageclassfication, liu2018patchobjectdetection, eykholt2018patchtrafficsign, sharif2016patchfacerecoginition}. Over time, adversarial attacks have expanded to domains including natural language processing~\cite{gao2018text}, speech recognition~\cite{carlini2018audio}, and autonomous driving~\cite{cao2019av}, underscoring the growing need for robust neural systems.

In embodied navigation, attacks have shifted toward more complex, interactive scenarios~\cite{xing2025embodiedaisafetysurvey, wang2025embodiednavigationsafetysurvey}. While some works exploit the vulnerabilities of VLN models through instruction-level destroying~\cite{lin2021insatt,lyu2025badnaver}, others focus on the perception-level attack.
Among vision-related attacks,
scene-centric methods manipulate the environment to mislead agents. For example, Spatiotemporal attack~\cite{liu2020spatiotemporal} generates 3D adversarial examples to alter object properties in key views, deceiving EQA agents, while \cite{yang2025hijacking} uses a differentiable renderer to optimize RGB textures of contextual objects, causing path deviations or premature termination. \cite{chen2024physicalizable} applies adversarial patches to specific scene objects, optimizing texture and opacity for physically realizable attacks in Object Goal Navigation. Agent-centric attacks directly manipulate the agent’s visual stream. Consistent Attack~\cite{ying2023uap} introduces temporally consistent universal perturbations across navigation trajectories, effectively degrading PPO-based agents’ performance. 
Despite these advances, most observation-level attacks assume white-box access to the target model gradient, which have limited applicability because of access restrictions in real-world deployment and substantial computational cost for gradient-based perturbation optimization. While black-box attacks~\cite{chen2017zoo, ilyas2018nes, guo2019simba} have been widely studied in vision tasks~\cite{bhambri2019blackboxCV}, existing methods lack a framework for continuously perturbing multi-step visual inputs and reliably disrupting multi-modal decision-making without model internal access. To address this gap, we propose a black-box attack framework that injects structured perturbations into the visual stream of the navigation agent to induce sustained trajectory deviations, extending black-box attack paradigms to VLN and exploring the development of its robustness.
\begin{figure*}[htbp]
    \centering
    \includegraphics[width=\linewidth]{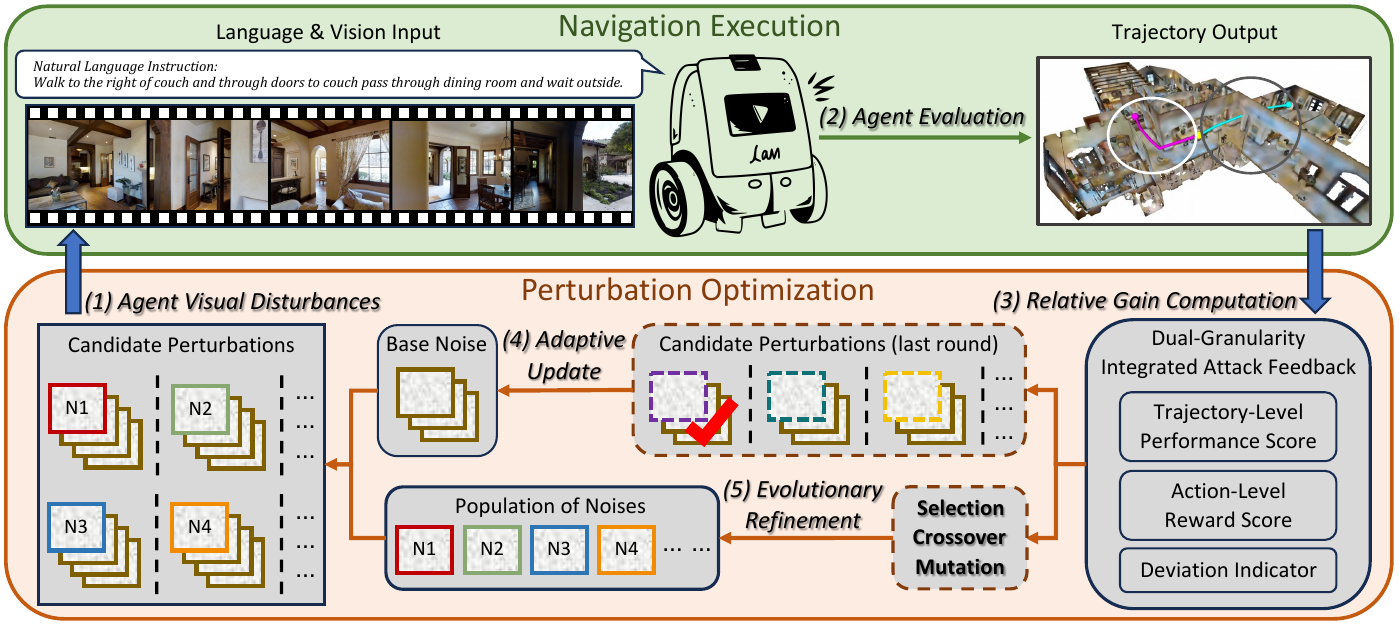}
    \caption{Behavior-guided perturbation optimization loop under black-box setting. (1) Generate trial candidate perturbations by sampling each noise from the population and overlaying it on the current base. (2) Evaluate the agent under each trial perturbation. (3) Compute attack relative gain using a dual-granularity feedback from trajectory outputs, including trajectory-level score, action-level score, and a deviation indicator. (4) Update the base if the gain improves. Otherwise, retain it and flip the search direction. (5) Evolve the population via selection, crossover, and mutation for the next round (see \Cref{ss: signals,ss: opt}).
    }
    \label{fig:framework}
\end{figure*}


\section{Black-Box Adversarial Attacks on VLN}
\label{s: methods}

In this section, we introduce a novel black-box adversarial attack framework for VLN task, as illustrated in \Cref{fig:framework}. 
We first propose a dual-granularity behavior-based feedback leveraging a trajectory-level performance score, an action-level reward score, and a deviation indicator as the optimization guidance in the black-box setting (see \Cref{ss: signals}). 
Based on this, we develop an adaptive optimization strategy that jointly heuristically tunes perturbation intensity and refines noise structure via genetic evolution (see \Cref{ss: opt}).

\subsection{Preliminaries}
\label{ss: preliminaries}

\noindent\textbf{VLN Task Definition.} In VLN, an agent is placed in an environment $\mathcal{E}$ and instructed with natural language $I$ to reach a target. At timestep $t$, the agent observes $v_t$ and retains history $ \mathcal{H}_t = \{(v_k,a_k)\}_{k=0}^{t-1}$. A policy $\pi$ selects an action 
\begin{equation}
    a_t = \pi(v_t,\mathcal{H}_t,I)\in\mathcal{A}_t,
\end{equation}
where the action space $\mathcal{A}_t$ includes a discrete \texttt{STOP} and movements to adjacent viewpoints on the scene graph. A navigation episode ends when \texttt{STOP} is issued, or a maximum action horizon $T$ is reached, producing a trajectory $\tau = (v_0,a_0,\ldots,v_T)$.

Navigation is evaluated with standard VLN metrics, including Success Rate (SR), Success weighted by Path Length (SPL), and Navigation Error (NE)~\cite{wu2024embodiednavigationsurvey}. Formally, for $N$ episodes:
\begin{itemize}
    \item SR (Success Rate) indicates the fraction of episodes whose final position lies within a threshold of the goal;
    \item SPL (Success weighted by Path Length) evaluates the path efficiency when the task is completed successfully;
    \item NE (Navigation Error) measures the Euclidean distance between the stopping point and the goal location.
\end{itemize}
These metrics are used in our optimization signal (see~\Cref{ss: signals}).

\noindent\textbf{VLN Model Architecture.} 
Transformer-based and LLM-based models employ distinct modes to process multimodal inputs and generate navigation actions.
For Transformer-based model, we choose HAMT~\cite{chen2021hamt} as our attack target, which at each step fuses the instruction, the current image, and the history of past observations and actions, followed by a cross-modal alignment, and then picks the next move from all candidate actions.
For LLM-based model, we choose MapGPT~\cite{chen2024mapgpt} as our attack target, which utilizes zero-shot large models to process multimodal input information and make movement decisions after reasoning.
Therefore, perturbations on the visual observations may distort their spatial understanding,
leading to subsequent wrong action selection and results in accumulated trajectory deviations and navigation failure.

\noindent\textbf{Black-box Adversarial Attack.} 
We consider the black-box adversarial attack access as query access~\cite{chen2017zoo, ilyas2018nes}, where the attacker is restricted to feeding modified visual observations to the agent and recording the output action possibilities for all valid actions with the final navigation metrics. This setting prohibits the attacker from computing and analyzing the gradient as in the white-box scenario.
Formulally, at step $t$, the clean view $v_t$ is transformed to
\begin{equation}
    \tilde v_t \;=\; \mathrm{clip}\big(v_t \;+\; \delta\big),
\end{equation}
where $\delta\in\mathbb{R}^{H\times W\times 3}$ is a unified perturbation applied to every frame with constraint $\|\delta\|_\infty\le\epsilon$. The perturbed trajectory is $\tau(\delta)=(\tilde v_0,\tilde a_0,\ldots,\tilde v_T)$. The attack seeks $\delta$ that maximizes navigation disruption under the budget constraint:

\begin{equation}
    \delta^\star
=\arg\max_{\ \|\delta\|_\infty\le \epsilon}\;
\mathcal{L}_{\text{attack}}\!\big(\tau(\delta);I,\mathcal{E}\big),
\label{eq: noise setting}
\end{equation}
where $\mathcal{L}_{\text{attack}}$ measures VLN performance degradation (our design detailed in~\Cref{ss: opt}).
In our method, $\delta$ is parameterized as a Perlin noise texture~\cite{perlin1985perlin}. This continuous representation produces smooth luminance and color variations that simulate disturbances such as fog and dust on the lens, and is controlled by a few parameters, facilitating efficient low-dimensional black-box search.

\subsection{Dual-Granularity Feedback Guidance for Behavior-based Attacks}
\label{ss: signals}

For the guidance of gradient-free optimization in the black-box setting, we design a behavior-driven dual-granularity attack feedback.
This feedback includes (i) a trajectory-level performance score $\mathcal{G}$ adapted from the standard VLN metrics quantifying the overall navigation performance degradation, (ii) an action-level reward score $\mathcal{R}$ capturing the potential step-wise decision drifts, and (iii) a binary indicator $\gamma$ explicitly flagging trajectory deviation.

\noindent\textbf{Trajectory-Level Performance Score.} 
Although internal model states are inaccessible, standard VLN metrics (SR, SPL, NE) remain observable and thus serve as behavioral feedback. An attack is successful if the agent completes the instruction under clean visual inputs but fails under perturbation, making SR a natural and direct indicator of attack success, \textit{i.e.}, the attack is more powerful when SR is lower. However, SR is binary for each instruction running result and provides little continuity for iterative optimization guidance. To construct a more informative signal from navigation evaluation results, we combine SPL and NE into a unified trajectory-level attack performance score:
\begin{equation}
    \mathcal{G} = \lambda \cdot \text{Norm}(\Delta\text{NE}) - (1 - \lambda) \cdot \text{Norm}(\Delta\text{SPL}), \quad \lambda \in [0, 1],
\end{equation}
where $\text{Norm}(\cdot)$ denotes a normalization factor to align the scales of NE and SPL, and $\Delta$ denotes the difference of the corresponding metric in the perturbed condition and in the clean one.

Maximizing $\mathcal{G}$ encourages the agent to deviate further from the goal location (higher $\Delta$NE) and reduce its trajectory efficiency (lower $\Delta$SPL), which indicates stronger navigation disruption and thus aligns with the attack objective.
However, due to some short Room-to-Room (R2R) instructions and the discrete Matterport3D environment, SPL and NE also exhibit near-binary changes in these situations, leading to guidance instability if used alone. This motivates a complementary fine-grained signal.

\noindent\textbf{Action-Level Reward Score.} While $\mathcal{G}$ summarizes the overall trajectory outcomes, it can also be sparse for some instructions. We therefore introduce an accumulated action-level attack reward score $\mathcal{R}$ that quantifies how perturbations erode the agent’s ability in choosing the correct action at each timestep. Even if the final trajectory remains unchanged under the perturbation, such "soft failures" captured by $\mathcal{R}$ reveal a latent tendency to choose the wrong action in the navigation process, increasing the risk of possible trajectory deviation. Formally, at step $t$,
\begin{equation}
    \mathcal{R} \;=\; \sum_{t=1}^{T}\!\Big(z_t^{\text{sec}} - z_t^{\text{opt}}\Big),
\end{equation}
where $z_t^{\text{opt}}$ is the possibility for choosing the reference action and $z_t^{\text{sec}}$ is that for the most likely incorrect alternative. Aggregation over all steps respects the sequential nature of VLN, where each decision possibly influences subsequent choices, thereby accumulating into long-term behavioral drift.

\noindent\textbf{Dual-Granularity Integrated Feedback Set.} 
We integrate the aforementioned signals into a unified attack feedback set for each candidate perturbation at round $r$ as
\begin{equation}
    \mathcal{S}_r\!\big(\widetilde{\eta}^{(r)}(\theta_n)\big)
\!=\! \Big\{\mathcal{G}\!\big(\widetilde{\eta}^{(r)}(\theta_n)\big),\;
             \mathcal{R}\!\big(\widetilde{\eta}^{(r)}(\theta_n)\big),\;
             \gamma\!\big(\widetilde{\eta}^{(r)}(\theta_n)\big)\Big\}.
\end{equation}

Specifically, for each candidate $\theta_n$, we generate $\delta(\theta_n)$ and update the previous base noise $\eta^{(r-1)}$ from the last round to form the trial candidate $\widetilde{\eta}^{(r)}(\theta_n)$ for this round, and evaluate the agent under this perturbation. Here, $\mathcal{G}$ captures trajectory-level disruption, $\mathcal{R}$ reflects action-level drift, and $\gamma$ indicates whether the agent actually deviates from the reference trajectory. 


\subsection{Adaptive Perturbation Optimization}
\label{ss: opt}

Building on the dual-granularity attack feedback introduced in~\cref{ss: signals}, we describe how it is operationalized to drive perturbation optimization in this section. At a high level, we maintain a cumulative base perturbation and, in each round:
(i) generate new candidate base noises by overlaying population-derived perturbation patterns onto the current base with a fixed step and accumulation sign,
(ii) evaluate candidates by their relative attack improvement on hierarchical signals (trajectory-level $\mathcal{G}$ or action-level $\mathcal{R}$, gated by $\gamma$), and
(iii) either update the base noise as the best candidate(if beneficial), or retain it while flipping the sign (if none are more effective), followed by population genetic evolutionary.

\noindent\textbf{Candidate Perturbation Generation.} 
Let $\eta^{(r-1)}$ denote the cumulative base perturbation obtained from the last round $r-1$. Each parameter vector $\theta$ in the noise population specifies a perturbation pattern $\delta(\theta)$. In round $r$, the trial provisional perturbation for candidate $\theta_n$ from the genetic pool is
\begin{equation}
    \widetilde{\eta}^{(r)}(\theta_n) \;=\; \eta^{(r-1)} \;+\; \alpha \cdot \text{sign} \cdot \delta(\theta_n),
\end{equation}
where $\alpha$ is the fixed step size, and $\text{sign}\!\in\!\{-1,+1\}$ denotes the current update direction, which indicates whether this sampled pattern is added to or subtracted from the current base. Applying $\widetilde{\eta}^{(r)}(\theta_n)$ across the each instruction entire execution process produces a trajectory $\tau\big(\delta(\theta_n)\big)$, from which we extract the dual-granularity integrated attack feedback set $\mathcal{S}_r\!\big(\widetilde{\eta}^{(r)}(\theta_n)\big)$.

\noindent\textbf{Relative Gain Computation.}
To assess each candidate's effectiveness, we compute relative attack gain with respect to the previous base $\eta^{(r-1)}$. For candidate $\theta_n$ in round $r$, the gain function is
\begin{equation}
    \mathcal{F}_r(\theta_n)
=
\begin{cases}
\mathcal{G}\!\big(\widetilde{\eta}^{(r)}(\theta_n)\big) - \mathcal{G}\!\big(\eta^{(r-1)}\big), \ \text{if } \gamma\!\big(\widetilde{\eta}^{(r)}(\theta_n)\big)=1, \\[4pt]
\mathcal{R}\!\big(\widetilde{\eta}^{(r)}(\theta_n)\big) - \mathcal{R}\!\big(\eta^{(r-1)}\big), \ \text{if } \gamma\!\big(\widetilde{\eta}^{(r)}(\theta_n)\big)=0.
\end{cases}
\end{equation}

\begin{figure}[t]
    \centering
    \includegraphics[width=0.85\linewidth]{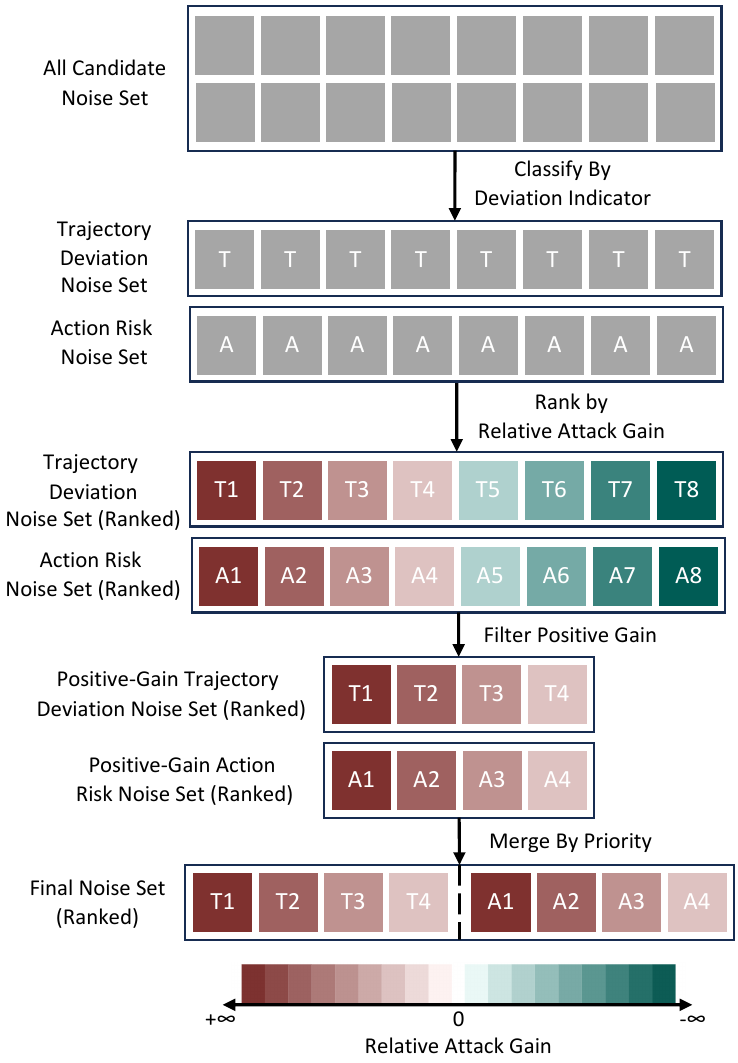}
    \caption{
    Ranking of trial candidate noises by relative gain. Candidates are first split by deviation indicator $\gamma$ into trajectory deviation vs. action risk sets. Within each set, ranking is by relative gain (warmer colors = larger positive gain; cooler = negative). Non‑positive gains are discarded. The two positive‑gain sets are then merged, prioritizing trajectory deviation candidates, to form the final ranked set.}
    \label{fig:rank}
\end{figure}

\noindent\textbf{Adaptive Perturbation Update.}
We rank the attack effect improvement brought by each candidate, according to the relative gain with a $\gamma$-aware rule that prioritizes perturbations which already induce trajectory deviation over those that only increase potential action drift risk, illustrated in \Cref{fig:rank}. Let
\begin{equation}
\mathcal{C}_{t}\!=\!\{\theta_n \!\mid\! \gamma(\widetilde{\eta}^{(r)}(\theta_n))\!=\!1\}, \
\mathcal{C}_{a}\!=\!\{\theta_n \!\mid\! \gamma(\widetilde{\eta}^{(r)}(\theta_n))\!=\!0\},
\end{equation}
and define the positive-gain subsets as
\begin{equation}
\mathcal{P}_{t}\!=\!\{\theta\!\in\!\mathcal{C}_{t}\!\mid\! \mathcal{F}_r(\theta_n)\!>\!0\}, \
\mathcal{P}_{a}\!=\!\{\theta\!\in\!\mathcal{C}_{a}\!\mid\! \mathcal{F}_r(\theta_n)\!>\!0\}.
\end{equation}
Then we select the best trial candidate which improves the attack effect the most, following a two-stage priority selection method as
\begin{equation}
\theta^*=
\begin{cases}
\arg\max_{\theta_n\in\mathcal{P}_{t}}\mathcal{F}_r(\theta_n), & \mathcal{P}_{t}\neq\varnothing,\\[8pt]
\arg\max_{\theta_n\in\mathcal{P}_{a}}\mathcal{F}_r(\theta_n), & \mathcal{P}_{t}=\varnothing\ \text{and}\ \mathcal{P}_{a}\neq\varnothing,\\[4pt]
\text{none}, & \mathcal{P}_{t}=\varnothing\ \text{and}\ \mathcal{P}_{a}=\varnothing.
\end{cases}
\end{equation}
If the best candidate $\theta^*$ exists, we update the current base noise to it. Otherwise, the base obtained from the last round is preserved, and the update direction is flipped, as
\begin{equation}
\eta^{(r)}=
\begin{cases}
\eta^{(r-1)}+\alpha\cdot \text{sign}\cdot \delta(\theta^*), & \theta^*\ \text{exists},\\[4pt]
\eta^{(r-1)},\quad \text{sign}\leftarrow-\text{sign}, & \text{otherwise}.
\end{cases}
\end{equation}

\noindent\textbf{Population Evolutionary Refinement.}
We reuse the same $\gamma$-aware candidate ranking result for population genetic operations each round, to refine possible accumulated noise structures.
The top-$k$ candidates are selected as parents, from which crossover and mutation generate the next generation. Deduplication and random resampling are then conducted to maintain population diversity. 


\begin{table*}[t]
\centering
\caption{Black-box attack performance of AdvNav and baseline methods on the Transformer-based VLN model (HAMT) across 11 scenes in the R2R val-unseen split. Metrics evaluated are Attack Success Rate (ASR $\uparrow$), Success weighted by Path Length (SPL $\downarrow$), and Navigation Error (NE $\uparrow$).
}
\label{tab:attack_results_by_scene}
\resizebox{0.9\linewidth}{!}{
\begin{tabular}{c|l|*{11}{c} | c}
\toprule
\multirow{2}{*}{\textbf{Method}} &
\multirow{2}{*}{\textbf{Metrics}} & 
\multicolumn{11}{c}{\textbf{Scene}} \\
& & \textbf{\MakeUppercase{\romannumeral 1}} & \textbf{\MakeUppercase{\romannumeral 2}} & \textbf{\MakeUppercase{\romannumeral 3}} & \textbf{\MakeUppercase{\romannumeral 4}} & \textbf{\MakeUppercase{\romannumeral 5}} & \textbf{\MakeUppercase{\romannumeral 6}} & \textbf{\MakeUppercase{\romannumeral 7}} & \textbf{\MakeUppercase{\romannumeral 8}} & \textbf{\MakeUppercase{\romannumeral 9}} & \textbf{\MakeUppercase{\romannumeral 10}} & \textbf{\MakeUppercase{\romannumeral 11}} & \textbf{Avg.}\\
\midrule
\multirowcell{3}{No Attack}
  & ASR$\uparrow$(\%) & 0.00 & 0.00 & 0.00 & 0.00 & 0.00 & 0.00 & 0.00 & 0.00 & 0.00 & 0.00 & 0.00 & 0.00 \\
  & SPL$\downarrow$(\%) & 50.64 & 46.76 & 62.17 & 62.69 & 30.66 & 53.91 & 59.34 & 69.69 & 50.29 & 50.53 & 57.96 & 57.69\\
  & NE$\uparrow$(m)     & 4.04 & 4.08 & 3.24 & 2.63 & 4.50 & 5.50 & 3.52 & 2.58 & 2.96 & 8.31 & 3.69 & 3.94\\
\midrule
\multirowcell{3}{Brightness Shift \\ (ZOO-tuned)~\cite{chen2017zoo}}
  & ASR$\uparrow$(\%) & 10.92 & 8.00 & 3.85 & 5.50 & 0.00 & 11.90 & 6.57 & 4.57 & 8.75 & 6.12 & 11.11 & 7.67\\
  & SPL$\downarrow$(\%) & 46.22 & 44.14 & 59.73 & 58.96 & 31.57 & 48.62 & 55.92 & 66.87 & 47.65 & 48.01 & 52.47 & 53.93\\
  & NE$\uparrow$(m)    & 4.34 & 4.31 & 3.56 & 2.80 & 4.43 & 6.01 & 3.75 & 2.77 & 3.14 & 8.79 & 4.13 & 4.25\\
\midrule
\multirowcell{3}{Mask Occlusion \\ (Bayes-tuned)~\cite{shukla2019bayesian}}
  & ASR$\uparrow$(\%) & 47.12 & \textbf{40.00} & \textbf{33.85} & 26.00 & 50.00 & 47.62 & 33.84 & 32.88 & 55.00 & 36.73 & 45.50 & 38.74\\
  & SPL$\downarrow$(\%) & 26.59 & 27.66 & \textbf{38.52} & 46.03 & 15.05 & 28.46 & 37.61 & 44.98 & 22.95 & 31.41 & 31.10 & 34.64\\
  & NE$\uparrow$(m)    & 5.90 & \textbf{6.12} & \textbf{5.27} & 3.51 & 5.42 & 7.58 & 4.96 & 4.23 & 4.65 & 10.77 & 5.67 & 5.68\\
\midrule
\multirowcell{3}{Gaussian Noise \\ (NES-tuned)~\cite{ilyas2018nes}}
  & ASR$\uparrow$(\%) & 28.16 & 24.00 & 10.00 & 13.00 & 33.33 & 26.19 & 15.15 & 15.53 & 17.50 & 19.39 & 24.87 & 19.10\\
  & SPL$\downarrow$(\%) & 38.18 & 36.95 & 54.98 & 55.11 & 21.56 & 41.12 & 51.55 & 58.81 & 43.12 & 41.47 & 43.88 & 47.48\\
  & NE$\uparrow$(m)     & 4.97 & 5.36 & 3.84 & 2.87 & 5.16 & 6.49 & 4.07 & 3.17 & 3.44 & 9.46 & 4.70 & 4.68\\
\midrule
\rowcolor{gray!20} 
  & ASR$\uparrow$(\%) & \textbf{52.87} & \textbf{40.00} & \textbf{33.85} & \textbf{45.50} & \textbf{100.00} & \textbf{55.36} & \textbf{40.40} & \textbf{43.84} &\textbf{67.50} & \textbf{41.84} & \textbf{69.84} & \textbf{49.70}\\
\rowcolor{gray!20} 
  & SPL$\downarrow$(\%) & \textbf{24.84} & \textbf{27.34} & 40.00 & \textbf{34.02} & \textbf{0.00} & \textbf{25.54} & \textbf{35.37} & \textbf{39.64} & \textbf{16.98} & \textbf{28.73} & \textbf{18.05} & \textbf{29.35}\\
\rowcolor{gray!20}[\tabcolsep] 
\multirow{-3}{*}{\textbf{AdvNav (ours)}}
  & NE$\uparrow$(m)    & \textbf{6.06} & 6.10 & 5.17 & \textbf{3.95} & \textbf{6.04} & \textbf{8.15} & \textbf{5.13} & \textbf{4.52} & \textbf{4.83} & \textbf{10.90} & \textbf{6.79} & \textbf{6.05}\\
\bottomrule
\end{tabular}
}
\end{table*}

\begin{figure*}[t]
    \centering
    \includegraphics[width=0.88\linewidth]{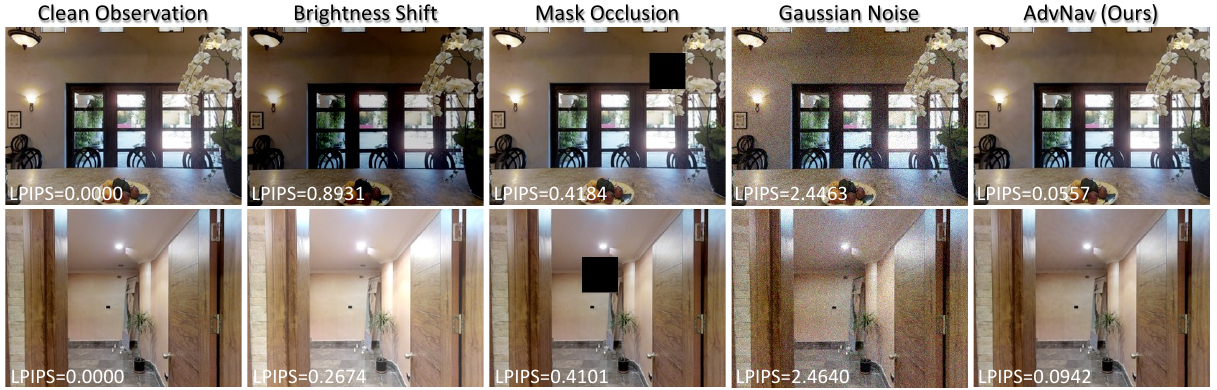}
    \caption{Visual comparison of the agent's first-person observations under different perturbations. We visualize how different adversarial strategies affect the agent’s visual input in two distinct scenes. From left to right: clean input, brightness shift, localized mask occlusion, Gaussian noise, and perturbation from our method. LPIPS score is annotated in the bottom left corner of each image (lower is better), quantifying the perceptual stealthiness of each perturbation.}
    \label{fig:resultvisual}
\end{figure*}
\section{Experiments}
\label{ss: experiments}

\subsection{Experimental Setup}
\label{ss: setup}

\textbf{Evaluation setting.} 
We evaluate our method on two VLN models, Transformer-based HAMT~\cite{chen2021hamt} and LLM-based MapGPT~\cite{chen2024mapgpt} using the Room-to-Room (R2R) dataset~\cite{anderson2018vlnr2r} within the Matterport3D (MP3D) simulator~\cite{Matterport3D}. 
All attacks are conducted on the \emph{val-unseen} split across 11 different indoor navigation scenes, encompassing 2349 instructions for HAMT and a randomly sampled subset of 165 instructions (15 per scene) for MapGPT.
We operate under the black-box setting with the query access, where the attacker has no access to model gradients and can only utilize observable input-output behavior.
For model configuration, we adopt the official R2R pretrained weights without any fine-tuning for HAMT, and set Qwen3-VL-32B-Instruct~\cite{Bai2025Qwen3VLTR} and GPT-4V~\cite{2023GPT4VisionSC} as the backbone for MapGPT, while both follow the original evaluation protocol.
By default, we run up to 20 rounds per instruction with a population size of 10 candidates, and set $\lambda$ 0.5 and $\alpha$ 1. We follow standard GA settings~\cite{lambora2019gareview} with population size 10, $k=2$, and mutation rate 0.2. 
All of our experiments are conducted on NVIDIA GeForce RTX 4080 SUPER 16GB.

\noindent\textbf{Baselines.} \textit{Prior VLN attack studies mostly assume white-box access or require model-specific training, making direct comparison infeasible.} To systematically evaluate AdvNav under the same black-box protocol and query budget, we design the following baselines:

\textit{No Attack:} No perturbation is applied; the agent performs navigation on clean visual observations. This serves as the reference for comparing pre-/post-attack metrics.

\textit{Brightness Shift (ZOO-tuned):} A smooth, unified pixel-wise intensity shift is applied to the visual input, simulating a global brightness attack. The perturbation magnitude is optimized via a black-box strategy akin to Zeroth-Order Optimization (ZOO)~\cite{chen2017zoo}.

\textit{Position-Optimized Mask Occlusion (Bayes-tuned):} A fixed-shape black patch simulates camera occlusion. Patch location is optimized using Bayesian search, while size and pixel value remain fixed~\cite{shukla2019bayesian}.

\textit{Gaussian Noise (NES-tuned):} Gaussian noise is added to the visual input and optimized via Natural Evolution Strategies (NES), tuning both mean and standard deviation~\cite{ilyas2018nes}.

\noindent\textbf{Evaluation Metrics.} We use Attack Success Rate (ASR) as the primary metric, measuring the proportion of instructions where the agent succeeds in the clean setting but fails under attack. To further quantify navigation degradation, we also report 
Success weighted by Path Length (SPL) and Navigation Error (NE).


\begin{figure*}[t]
    \centering
    \includegraphics[width=0.90\linewidth]{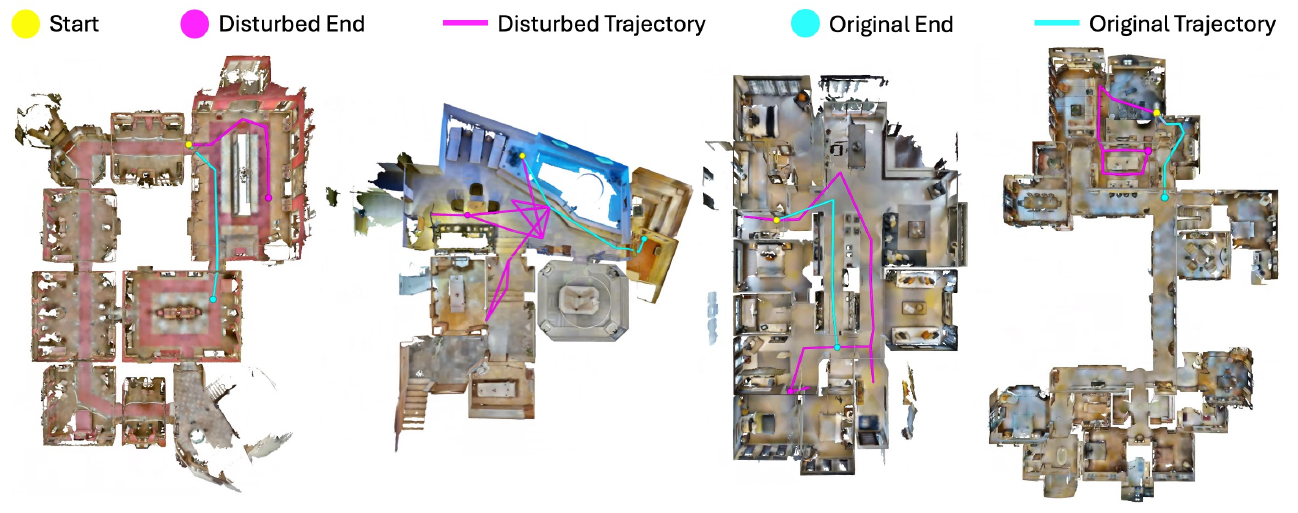}
    \caption{Trajectory deviation induced by adversarial perturbations. We visualize the agent’s navigation trajectories before and after our attack on four different scenes. 
    Each map overlays the original trajectory (cyan line) and its endpoint (cyan dot) with the perturbed trajectory (pink line) and disturbed endpoint (pink dot), both originating from the same start position (yellow). Under attack, the agent consistently deviates from its intended trajectory and finally fails to reach the goal.}
    \label{fig:traj_result}
\end{figure*}


\subsection{Black-Box Attack Performance on Transformer-Based VLN (HAMT)}
\label{ss: hamt_result}

\Cref{tab:attack_results_by_scene} presents scene-wise and average attack performance of AdvNav and all baselines against Transformer-based HAMT under the black-box setting. 
On average, AdvNav achieves ASR 49.70\%, SPL 29.35\%, and NE 6.05 m. Compared to the clean input (SPL 57.69\%, NE 3.94 m), SPL drops by 28.34\% and NE increases by 2.11 m. Relative to the strongest baseline, Mask Occlusion (ASR 38.74\%, SPL 34.64\%, NE 5.68 m), AdvNav approximately improves ASR by 11\%, reduces SPL by 5\%, and increases NE by 0.4 m. Gains over Gaussian Noise and Brightness Shift are even larger across all metrics. This comparison demonstrates that AdvNav effectively identifies and exploits visual vulnerabilities, and substantially disrupts the VLN agent's perception and navigation performance.
For the detailed scene-wise analysis, AdvNav consistently ranks first or ties for first across all 11 scenes. Scene-wise ASR generally ranges from 40\% to 70\%, with a maximum of 100\% in scene \MakeUppercase{\romannumeral 5}, leading to complete task failure. Relative to Mask Occlusion, ASR improvements range from 5.11\% (scene \MakeUppercase{\romannumeral 10}) to 50.00\% (scene \MakeUppercase{\romannumeral 5}). SPL drops, and NE increases accompany most scenes, indicating stronger trajectory deviation. Compared with Gaussian Noise and Brightness Shift, AdvNav achieves higher ASR in every scene, typically paired with lower SPL and higher NE. These results mean that AdvNav demonstrates stable
black-box attack performance across diverse layouts and instructions without scenario-specific prerequisites.
For cost, our average attack round is 6.77, with a time spent of about 32 seconds each round, occupying about 2.4 GB of GPU memory.

\Cref{fig:resultvisual} compares how different perturbations alter the agent's first-person observations. For quantitative evaluation, we introduce LPIPS (Learned Perceptual Image Patch Similarity)~\cite{zhang2018lpips} to measure the perceptual distance between the perturbed and clean observations. A score closer to zero indicates that the perturbation is highly imperceptible and visually similar to the original image. Our AdvNav achieves a significantly lower LPIPS score compared to other perturbations, demonstrating its perceptual stealthiness while maintaining strong attack efficacy. 
For qualitative analysis, brightness Shift simulates global illumination changes, and its spatial uniformity preserves contrast and geometry, while normalization may suppress its effects. Mask Occlusion hides a local region, strongly disrupting perception when covering relevant content, but its impact is highly viewpoint-dependent. Gaussian Noise introduces fine-grained fluctuations resembling sensor noise, where edges and structures remain intact, and downsampling or denoising may alleviate the disturbance.
In contrast, our perturbation forms a low-frequency, coherent luminance field (akin to haze or dust on the lens). It reduces contrast along structural cues and resists suppression by standard preprocessing, leading to continuous bias in navigation action that accumulates across trajectories. 
\Cref{fig:traj_result} clearly shows the obvious trajectory deviation caused by our method's perturbation in different scenes.

\subsection{Black-Box Attack Performance on LLM-Based VLN (MapGPT)}
\label{ss: mapgpt_result}

As illustrated in Table~\ref{tab:attack_result_mapgpt}, AdvNav achieves a potent adversarial impact on both backbones of zero-shot LLM-based MapGPT. Specifically, AdvNav achieves 65.96\% ASR on Qwen3-VL backbone, with a 43.12\% drop in SPL and 2.35 m increase in NE, while an ASR of 87.30\% on GPT-4V backbone, reducing SPL to a mere 4.71\% with 2.04 m growth in NE, revealing the visual vulnerability of advanced LLM-based navigation agents to our perturbations. 
Another intriguing finding is the shift in sensitivity to different perturbation types. While Mask Occlusion, which only affects partial visual loss, is highly effective against HAMT, it proves less impactful on MapGPT. Conversely, Gaussian Noise and Brightness Shift, which affect the holistic observation, induce much stronger attack performance in MapGPT than HAMT. Such distinct trends suggest a potential difference in the visual robustness bottleneck between the two types of models. 
However, the strong attack performance of our strategy across both HAMT and MapGPT underscores its effectiveness and broad applicability, which demonstrates that AdvNav can serve as a general stress-testing framework, effectively identifying visual vulnerabilities regardless of the VLN agent's underlying visual perception and decision-making paradigm.

\begin{table}[t]
\centering
\caption{Black-box attack performance of AdvNav and baseline methods on the LLM-based VLN model (MapGPT) across two backbones.
}
\label{tab:attack_result_mapgpt}
\setlength{\tabcolsep}{1pt}
\resizebox{1.0\linewidth}{!}{
\begin{tabular}{c|ccc|ccc}
\toprule
\multirow{2}{*}{\textbf{Method}}
& \multicolumn{3}{c|}{\textbf{Qwen3-VL}} & \multicolumn{3}{c}{\textbf{GPT-4V}} \\
\cmidrule(lr){2-4} \cmidrule(lr){5-7} 
& ASR$\uparrow$(\%) & SPL$\downarrow$(\%) & NE$\uparrow$(m) 
& ASR$\uparrow$(\%) & SPL$\downarrow$(\%) & NE$\uparrow$(m) \\
\midrule
No Attack & 0.00 & 61.12 & 3.82 & 0.00 & 39.91 & 5.35\\
{\small Brightness Shift (ZOO-tuned)~\cite{chen2017zoo}} & 48.60 & 31.25 & 5.41 & 55.26 & 18.87 & 7.05\\
{\small Mask Occlusion (Bayes-tuned)~\cite{shukla2019bayesian}} & 17.78 & 41.64 & 4.82 & 27.40 & 28.08 & 6.09\\
{\small Gaussian Noise (NES-tuned)~\cite{ilyas2018nes}} & 58.76 & 21.71 & 6.09 & 76.92 & 10.45 & \textbf{7.41}\\
\rowcolor{gray!20} 
\textbf{AdvNav(ours)} & \textbf{65.96} & \textbf{18.00} & \textbf{6.17} & \textbf{87.30} & \textbf{4.71} & 7.39\\
\bottomrule
\end{tabular}}
\end{table}

\subsection{Sensitivity Discussion}
\label{ss: parameter}

\noindent{\textbf{Optimization parameter sensitivity analysis.}}
The tradeoff between performance and budget for AdvNav is dictated by the number of optimization rounds and noise population size.
To identify the optimal configuration, we conduct a grid search using HAMT on a subset of randomly chosen 165 instructions (15 per scene) from R2R val-unseen.
As shown in Table~\ref{tab:param_analysis}, ASR generally exhibits a positive correlation with both the number of rounds and the population size. 
When the population size is small, increasing the rounds yields limited improvements, as the low diversity of noise patterns restricts the search space. 
Similarly, at a low round count, expanding the population only results in marginal gains, demonstrating that the benefits of a larger population are only fully realized when given sufficient iterations to evolve.
We observe that the attack performance reaches a plateau at 20 rounds and a population size of 10, achieving an ASR of 47.19\%. Since further increasing the rounds or sizes results in no additional gains, we select 20 rounds and a population size of 10 as our default setting to balance attack performance with computational efficiency.

\noindent{\textbf{Noise structure sensitivity analysis.}}
Perlin noise generates spatially coherent luminance fields, where the noise structural configuration may influence the attack's effect. 
Therefore, we conduct experiments to investigate the attack sensitivity to noise structure. 
Specifically, we randomly generate five noises $\phi_a-\phi_e$ from the Perlin family. For each noise, the structure is fixed during the optimization while adaptive update guided by behavioral feedback is retained. 
We randomly choose 165 instructions (15 per scene) from R2R val-unseen split for this evaluation, and set HAMT as our target model.
\Cref{tab:noise_structure_results} shows that attack performance varies with structure: ASR ranges from 26.97\% to 33.71\%, SPL from 36.86\% to 32.94\%, and NE from 5.31 m to 5.74 m, and $\phi_e$ achieves the best performance (ASR 33.71\%, SPL 32.94\%, NE 5.74 m),
which motivates the use of structure genetic evolution in our full method.

\begin{table}[t]
\centering
\caption{Attack performance of AdvNav under different combinations of optimization round and population size. 
}
\label{tab:param_analysis}
\resizebox{0.7\linewidth}{!}{
\begin{tabular}{c|c|cccc}
\toprule
\multicolumn{2}{c|}{\multirow{2}{*}{\textbf{ASR↑(\%)}}} & \multicolumn{4}{c}{\textbf{Population Size}} \\
\cmidrule(lr){3-6}
\multicolumn{2}{c|}{} & 6 & 8 & 10 & 12 \\
\midrule
\multirow{4}{*}{\textbf{Round}} 
 & 10 & 35.96 & 37.08 & 38.20 & 38.20 \\
 & 15 & 37.08 & 38.20 & 41.57 & 41.57\\
 & 20 & 34.83 & 39.33 & \textbf{47.19} & 42.70\\
 & 25 & 38.20 & 41.57 & \textbf{47.19} & \textbf{47.19}\\
\bottomrule
\end{tabular}}
\end{table}

\begin{table}[t]
\centering
\caption{Attack performance of AdvNav under diverse randomly chosen noise spatial structures. 
}
\label{tab:noise_structure_results}
\resizebox{0.65\linewidth}{!}{
\begin{tabular}{c|ccc}
\toprule
\textbf{Noise} & \textbf{ASR$\uparrow$(\%)} & \textbf{SPL$\downarrow$(\%)} & \textbf{NE$\uparrow$(m)} \\
\midrule
$\phi_{a}$ & 26.97 & 36.86 & 5.31 \\
$\phi_{b}$ & 31.46 & 34.41 & 5.66 \\
$\phi_{c}$ & 32.58 & 34.52 & 5.72  \\
$\phi_{d}$ & 30.34 & 35.48 & 5.61 \\
$\phi_{e}$ & \textbf{33.71} & \textbf{32.94} & \textbf{5.74} \\
\bottomrule
\end{tabular}}
\end{table}

\subsection{Ablation Study}
\label{ss: ablation study}
We analyze the contributions of \emph{feedback guidance}, \emph{adaptive perturbation update}, and \emph{noise structure evolution} to our method. Experiment results are summarized in \Cref{tab:ablation_components}.

\noindent\textbf{Effect of Feedback Guidance.}
We investigate the contribution of our dual-granularity feedback and its individual components, including the trajectory-level performance score and the action-level reward score. As illustrated in the first three rows of \Cref{tab:ablation_components}, removing both feedback signals (pure random search) yields the weakest performance (23.40\% ASR), establishing the necessity of behavioral guidance for perturbation optimization. Specifically, employing only the trajectory-level score provides a moderate improvement to 30.85\% ASR, demonstrating that its sparsity limits its ability to guide optimization effectively. Utilizing the action-level score achieves a higher 45.74\% ASR, which confirms the effect of this more informative signal on guidance. However, the difference compared to our full feedback design (-17.02\% and -2.13\% respectively) illustrates that the synergy of dual-granularity signals is essential for achieving the most potent and stable adversarial impact.

\noindent\textbf{Effect of Adaptive Perturbation Update.} With structure evolution disabled and the best template $\phi_e$ fixed (see \Cref{ss: parameter}), adaptive update improves ASR by 14.90\%, reduces SPL by 7.35\%, and increases NE by 0.80 m over perturbation obtained by pure random search. This demonstrates that even a fixed structure benefits from the guided adaptive accumulation. However, ASR remains 9.57\% below the full method, showing that coupling with structure evolution is essential for maximal performance.

\noindent\textbf{Effect of Noise Structure Evolution.} 
Retaining structure evolution but disabling adaptive update produces 3.20\% higher ASR than pure random search, demonstrating that structure evolution alone prunes ineffective patterns and discovers task-relevant ones, providing moderate gains. Yet, without adaptive intensity update, the disruption of these optimized perturbations remains highly limited, yielding 21.27\% lower ASR than the full method.

\noindent\textbf{Effect of Overall Optimization Strategy.} 
We further compare against Perlin noise optimized with NES, Bayesian, and Simba methods, achieving ASRs of 39.33\%, 19.10\% and 20.22\%, respectively. These results show that the performance gains mainly come from our framework design rather than Perlin perturbations alone.


\begin{table}[t]
\centering
\caption{Ablation study on feedback guidance, adaptive perturbation update, and noise structure evolution. 
}
\label{tab:ablation_components}
\setlength{\tabcolsep}{1.5pt}
\resizebox{1.0\linewidth}{!}{
\begin{tabular}{c|c|cc|c|c|c}
\toprule
\textbf{Adaptive} & \textbf{Structure}
& \multicolumn{2}{c|}{\textbf{Feedback}} &
\multirow{2}{*}{\textbf{ASR$\uparrow$(\%)}}  &
\multirow{2}{*}{\textbf{SPL$\downarrow$(\%)}} &
\multirow{2}{*}{\textbf{NE$\uparrow$(m)}} \\ 
\cmidrule(lr){3-4} 
\textbf{Update} & \textbf{Evolution}
& \textbf{Trajectory} & \textbf{Action} & & &\\
\midrule
$\times$    & $\times$   & $\times$  & $\times$  & 23.40 & 39.39 & 4.54 \\
\checkmark  & \checkmark & \checkmark  & $\times$ & 30.85 & 36.48 & 4.77 \\
\checkmark  & \checkmark & $\times$  & \checkmark & 45.74 & 29.43 & 5.26\\
\cmidrule{1-7}
\checkmark  & $\times$  & \checkmark   & \checkmark    & 38.30 & 32.04 & 5.34 \\
$\times$    &\checkmark  & \checkmark   & \checkmark   & 26.60 & 38.77 & 4.56 \\
\checkmark  & \checkmark & \checkmark  & \checkmark   & \textbf{47.87} & \textbf{25.95} & \textbf{5.52} \\
\bottomrule
\end{tabular}}
\end{table}


\begin{table}[t]
\centering
\caption{Ablation study on our overall optimization. 
}
\label{tab:ablation_overall}
\resizebox{0.9\linewidth}{!}{
\begin{tabular}{c|ccc}
\toprule
\textbf{Optimization} & \textbf{ASR$\uparrow$(\%)} & \textbf{SPL$\downarrow$(\%)} & \textbf{NE$\uparrow$(m)} \\
\midrule
Perlin-NES~\cite{ilyas2018nes} & 39.33 & 29.88 & 6.04 \\
Perlin-Bayes~\cite{shukla2019bayesian} & 19.10 & 39.96 & 5.21 \\
Perlin-Simba~\cite{guo2019simba} & 20.22 & 39.41 &  5.56 \\
AdvNav & \textbf{47.87} & \textbf{25.95} & \textbf{5.52} \\
\bottomrule
\end{tabular}}
\end{table}

\section{Conclusion}

In this work, we introduce AdvNav, a black-box adversarial attack framework for VLN that utilizes visual disturbances optimized through perturbation adaptive intensity tuning and noise structure genetic evolution under the guidance of a dual-granularity behavior-based feedback. 
Our method achieves effective attacks against Transformer-based HAMT and LLM-based MapGPT on R2R val-unseen dataset with a limited query budget, 
thereby exposing the universal latent visual vulnerability of VLN models with varied architectures.
We position AdvNav not only as an attack method or stress-testing tool for VLN systems but also as a catalyst for future robustness research,
such as adversarial training.



\section{Acknowledgements}
This work was supported by National Natural Science Foundation of China (NFSC) under the Grant Number 62573370 and Key Area Project of Education Department of Guangdong Province (No. 2025ZDZX3051).

\bibliographystyle{ACM-Reference-Format}
\bibliography{acmmm26}

\end{document}